\pdfoutput=1

\documentclass[11pt]{article}

\usepackage{acl}
\usepackage{amsfonts}
\usepackage{subcaption}
\usepackage{listings}
\usepackage[ruled]{algorithm2e}

\usepackage{enumitem}

\usepackage{times}
\usepackage{booktabs}

\usepackage{latexsym}

\usepackage[T1]{fontenc}

\usepackage[utf8]{inputenc}

\usepackage{microtype}

\usepackage{graphicx}
\usepackage{amsmath}
\usepackage{cleveref}

\newcommand{\QS}{\mathit{{QualityScore}}}
\newcommand{\CS}{\mathit{ControlError}}
\newcommand{\CSs}{\mathit{CtrlError}}
\newcommand{\Cost}{\mathit{Cost}}
\newcommand{\Calm}{{\textsc{CaLM}}}
\definecolor{avtgreen}{RGB}{0, 150, 85}

\def\colorful{1}

\ifnum\colorful=1

\fi
\ifnum\colorful=0

\fi

\title{From Tarzan to Tolkien: Controlling the Language Proficiency Level of LLMs for Content Generation}

\author{
    Ali Malik \\
    Stanford University \\
    Stanford, CA \\
    \And
    Stephen Mayhew \\
    Duolingo \\
    Pittsburgh, PA \\
    \And
    Chris Piech \\
    Stanford University \\
    Stanford, CA \\
    \And
    Klinton Bicknell \\
    Duolingo \\
    Pittsburgh, PA \\
}

\begin{document}
\maketitle


\begin{abstract}

We study the problem of controlling the difficulty level of text generated by Large Language Models (LLMs) for contexts where end-users are not fully proficient, such as language learners. Using a novel framework, we evaluate the effectiveness of several key approaches for this task, including few-shot prompting, supervised finetuning, and reinforcement learning (RL), utilising both GPT-4 and open source alternatives like LLama2-7B and Mistral-7B.

Our findings reveal a large performance gap between GPT-4 and the open source models when using prompt-based strategies. However, we show how to bridge this gap with a careful combination of finetuning and RL alignment. Our best model, \Calm{} (CEFR-Aligned Language Model), surpasses the performance of GPT-4 and other strategies, at only a fraction of the cost. We further validate the quality of our results through a small-scale human study.

\end{abstract}

\section{Introduction}

\begin{figure}
    \centering
    \includegraphics[width=0.65\linewidth]{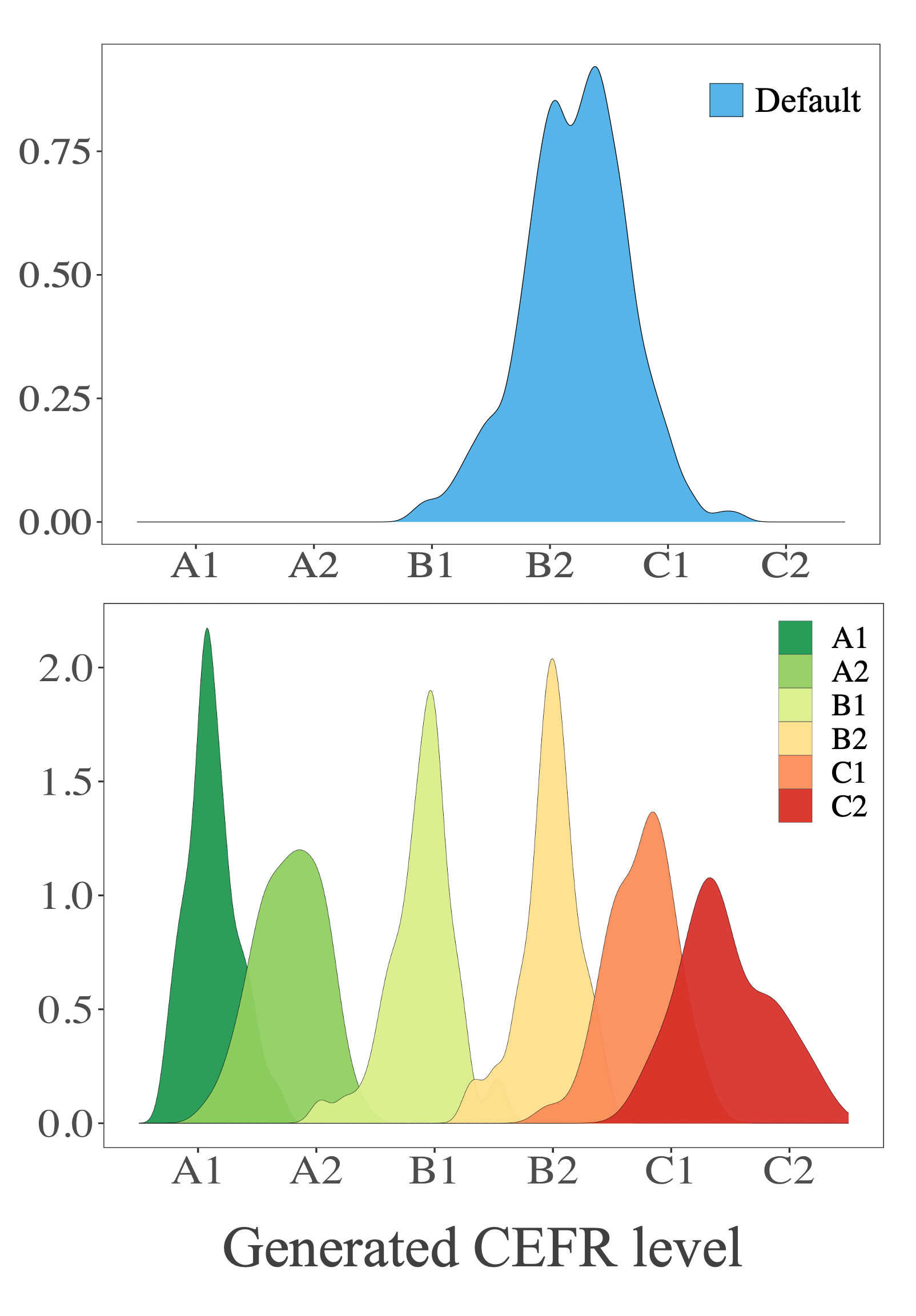}
    \caption{\textbf{(top)} GPT-4 generates content at a native proficiency level. 
    \textbf{(bottom)} Results from our \Calm{} proficiency control model for different target levels.}
    %
    
    \label{fig:uncond_level}
\end{figure}

Large Language Models (LLMs) are powerful tools for content generation. However, these models often output text at a native  level of speech (\Cref{fig:uncond_level}, top).
This makes LLMs challenging to use for contexts where the end users are not fully proficient, such as for language learners, young children, or non-native speakers. 
When generating content for these use cases, we need the ability to control the proficiency level of the generated text. 

In this work, we formally define the Proficiency Control Task (PCT): a new framework that assesses a model's ability to modulate language proficiency level, while also generating high-quality content consistent with given instructions. We evaluate models with respect to the three essential criteria: (1) $\CS$ which measures how close the generated text is to the target proficiency, (2) $\QS$, which measures relevance of the text to the instructions, and (3) $\Cost$ which measures the resource-intensiveness of the approach. 

Using this evaluation framework and the TinyStories dataset \cite{eldan2023tinystories}, we investigate several key approaches to the PCT on the task of short story generation from a plot summary.

\paragraph{Prompt-based approaches} First, we thoroughly explore the space of few-shot, prompt based strategies with OpenAI's GPT-4 and open source alternatives (\Cref{sec:prompt}). Our findings demonstrate the strong capability of GPT-4 at the PCT, resulting in low $\CS$ and high $\QS$. We also identify an improvement in $\CS$ as  prompts are made more complex, resulting in better proficiency control at the cost of more tokens.  

Although GPT-4 is successful at the PCT, it is a proprietary model and its generations are several times more costly than open source alternatives. However, we find instruction-tuned models like LLama-2-7b and Mistral-7b perform poorly at the PCT through prompting.

\paragraph{Finetuning open source models}

To bridge the gap between open source models and GPT-4, we turn to supervised finetuning approaches from the controllable text generation literature \cite{keskarCTRL2019,stowe-etal-2022-controlled}. Specifically, we use the outputs of an effective GPT4 prompting strategy to generate data for the PCT that can be used to directly train open source models.

Using this data, we are able to finetune LLaMa2-7b and Mistral-7b to come significantly closer in performance to GPT-4 at the PCT (\Cref{sec:ft}). Moreover, we show how additional training with Proximal Policy Optimisation (PPO) can further align the outputs of these models with the desired proficiency levels. Our best such model, we call \Calm{} (CEFR-Aligned Language Model), has a $\CS$ equal to that of GPT-4, at only a fraction of the cost.

\paragraph{Boosting PCT models through sampling}
Finally, in \Cref{sec:boosting} we present a simple but powerful sampling strategy that allows us to boost any PCT model to one with arbitrarily better $\CS$, albeit at a higher cost.  With this technique, we are able to show that \textbf{\textit{\Calm{} is a strictly dominant strategy}} in the Pareto sense compared to GPT-4 with \textit{any} kind of prompting.

We run a small-scale human evaluation (\Cref{sec:human-eval}) to further validate the quality of generations from \Calm{} and GPT-4 with prompting. The generations of both models are highly rated in terms of quality ($\approx 4.7$ out of 5). We also show that our measure of $\CS$ aligns closely with human perceptions of ``proficiency level''.

\section{Background: CEFR}
To discuss language proficiency levels, we employ the widely-used Common European Framework of Reference (CEFR) \cite{councilofeurope2001}.  The CEFR is a general framework that organises proficiency in any language into six levels of increasing proficiency: A1, A2, B1, B2, C1, C2, each defined through ‘can-do’ descriptors (\Cref{table:cefr_levels}).
The advantage of CEFR is that it is well-known in practice, allowing us to leverage existing expert-labelled datasets to create an automatic scorer.


\subsection{Automatic CEFR Scoring}\label{sec:cefr_scoring}
 

For our work, we need the ability to automatically score text proficiency. There is a long line of research on automated assessment for text readability \cite{schwarm-ostendorf-2005-reading,xia-etal-2016-text, pv18investigating}. We build upon this literature, but treat scoring as a regression problem, with $\{1, \ldots, 6\}$ corresponding to levels A1 through C2. We train a standard linear regression model with linguistic features using public datasets of human-labelled CEFR English texts \cite{xia-etal-2016-text, montgomerieCEFRdata, Breuker2023} (see \Cref{appx:scorer_details} for more details). Our scoring function demonstrates an $R^2$ on $0.8$ on a held-out test set. Moreover, in a human evaluation (\Cref{sec:human-eval}), we show that our scorer seems to align well with human perceptions of text proficiency. 

Due to the inherent ambiguity in CEFR descriptions and differing labelling criteria used across datasets, there is some arbitrariness in one's choice of automated CEFR scorer. While we use a particular scoring function in this work, all of the results presented in this paper use this function as a black box, allowing it to be modularly replaced with a different scorer as needed. We believe our results would generalise to any reasonable scoring function (see \Cref{appx:scorer_disc} for a discussion).





\section{The Proficiency Control Task}

We now formally define the Proficiency Control Task (PCT), which measures a model's ability to generate content relevant for a given prompt while also controlling the proficiency level of its output.

Formally, let $\Sigma^*$ denote the set of strings. Let $p \in \Sigma^*$ be a prompt and $t \in \{1,2,3,4,5,6\}$ be a target proficiency (corresponding to each CEFR level). We denote a Proficiency Control model as a function $\mathcal{M} : (\Sigma^* \times \{1,2,3,4,5,6\}) \to \Sigma^*$ that takes a  prompt and target proficiency as input and outputs a generated text for the given prompt.
We assess the PCT on three key criteria:


\paragraph{Control} This evaluates how close the generated text was to the target proficiency level.  Let $s_{\mathrm{cefr}} : \Sigma^* \to \mathbb{R}$ be an automatic proficiency scoring function.
We define the $\CS$ between a target proficiency $t$ and a generated text $x \in \Sigma^*$ as 
\begin{equation*}\label{eq:control_score}
    \CS(x, t) = (s_{\mathrm{cefr}}(x) - t)^2
\end{equation*}




\paragraph{Quality} This measures the relevance and quality of the generated content to the given prompt. For example, if the prompt asks for an English story with a certain plot, then the text should be in correct English and closely align with the given plot.

\paragraph{Cost} This measures how expensive the control strategy is with respect to various resources e.g. flops, time, or dollars. Our primary resource of interest for LLMs will be FLOPs, which are a function of the size of the model and the number of tokens used by the strategy.

\section{Strategies for Proficiency Control }

In this section we discuss several approaches to proficiency control for LLMs.  These approaches are broadly categorised into prompt-based techniques, supervised finetuning on a PCT dataset, and a general  sampling strategy to improve any PCT model.

\subsection{Prompt-based approaches}\label{ssec:prompt_strategies}

One of the simplest forms of eliciting desired behaviour from LLMs is through clever prompting. This approach is quick, easy to iterate, and can be used with the most powerful proprietary models. We explore different ways to construct prompts to control proficiency level. Each approach builds up in complexity by providing more useful information about the desired proficiency level, but at the cost of using more tokens. The full prompts for each strategy can be found in \Cref{appx:prompts}.

\paragraph{Baseline}
The simplest step to controlling proficiency is to directly ask the LLM to generate at a certain CEFR level (\textbf{Base}). Since LLMs are trained on massive amounts of data, they possess context about CEFR. For example, GPT-4 can produce an accurate description of each CEFR level. By prompting the model to generate at a level, it can draw on its existing knowledge to guide generation.

\paragraph{Describing CEFR}
The next improvement over the baseline strategy is to include concrete descriptions of the CEFR levels in the prompt. Here we can choose between describing just the target level (\textbf{Descr. (target)}) or describing every single CEFR level (\textbf{Descr. (all)}). The latter contains more information but the former is more efficient in terms of number of tokens used. We use official descriptions of the levels from the Council of Europe, which is the establishing body of CEFR.


\paragraph{Few-shot Learning} 

Several recent results have shown the power of including examples in the prompt to improve LLM generation \cite{rag_lewis_2020}. 
In the context of proficiency control, we can augment the descriptions of the CEFR levels with an expert-written example text at that level. As before, we can choose to include an example for only the target level (\textbf{Few (target)}) or for all CEFR levels (\textbf{Few (all)}).

\begin{table*}[t]
\small
\centering
\begin{tabular}{l|c|cc|cc|cc}
\toprule
 && \multicolumn{2}{c|}{\textbf{GPT4} (>175B)} & \multicolumn{2}{c|}{\textbf{LLama-2-7b-chat} (7B)} & \multicolumn{2}{c}{\textbf{Mistral-7b-Instruct} (7B)} \\
\textbf{Prompt Strategy} &  \# tokens $\downarrow$ & $\CSs$ $\downarrow$& Quality $\uparrow$ & $\CSs$ $\downarrow$ & Quality $\uparrow$ & $\CSs$ $\downarrow$ & Quality  $\uparrow$  \\
\midrule
\textbf{(-)} {Original} & 109 & 
$3.66 \pm 0.22$ & (9.5, 10) &
$3.23 \pm 0.17$ & (9.5, 9.7) & 
$3.89 \pm 0.23$ & (9.6, 9.9)   \\
\\
\textbf{(a)} {Base} & 132 &
    ${0.57 \pm 0.05}$ & (9.5, 10) & 
    $2.76 \pm 0.16$ & (9.4, 9.7) & 
    $1.68 \pm 0.10$ & (9.6, 9.9)  \\
\\
\textbf{(b)} {Descr. (target)} & 211&
    ${0.39 \pm 0.03}$ & (9.4, 10) &  
    $1.84 \pm 0.12$ & (9.4, 9.8) & 
    $1.20 \pm 0.08$ & (9.4, 9.9)  \\ 
\textbf{(c)} {+ Few (target)} & 458&
    ${0.39 \pm 0.03}$ & (9.4, 9.9) & 
    $2.05 \pm 0.13$ & (9.3, 9.7) &
    $1.30 \pm 0.08$ & (9.6, 9.9)  \\ 
\\
\textbf{(d)} {Descr. (all)} & 609 &
    ${0.34 \pm 0.03}$ & (9.4, 9.9) & 
    $1.67 \pm 0.10$ & (9.5, 9.7)  &
    $1.31 \pm 0.09$ & (9.4, 9.9)  \\ 
\textbf{(e)} {+ Few (target)} & 935 &
    ${0.28 \pm 0.03}$ & (9.4, 9.9) & 
    $1.53 \pm 0.10$ & (9.3, 9.6)  &
    $1.19 \pm 0.08$ & (9.5, 9.9)  \\ 
\textbf{(f)} {+ Few (all)} & 2206 &
    ${0.30 \pm 0.02}$ & (9.4, 9.9) &   
    $1.86 \pm 0.12$ & (9.4, 9.6)  & 
    $1.58 \pm 0.10$ & (9.6, 9.9)  \\ 
\bottomrule
\end{tabular}
\caption{Results of different prompting strategies on the TinyStories Proficiency Control Task. Quality scores are given as a tuple of (Fluency, Consistency) scores. The Cost value for each approach is proportional to the number of tokens for that strategy multiplied by the number of parameters of the model (shown in the Table heading).}
\label{tab:prompt_results}
\end{table*}

\begin{table*}
    \centering
    \small
  \begin{tabular}{lccc}
  \toprule
  \textbf{Model}  & $\CSs$ $\downarrow$ & Quality $\uparrow$ & \# tokens $\downarrow$ \\
  \midrule
    {Mistral-7b: Finetuned}  & $0.69 \pm 0.05$ & (9.4, 9.9) & 110  \\
    {Mistral-7b: Finetuned + PPO}  & $0.60 \pm 0.05$ & (9.1, 9.7) & 110  \\
    \\
    {LLama2-7b: Finetuned}  & $0.81 \pm 0.06$ & (9.3, 9.8) & 110  \\
    {\Calm{}: LLama2-7b Finetuned + PPO}  & ${0.39 \pm 0.03}$ & {(9.2, 9.6)} & 110  \\
    \\
    {\textbf{\Calm{} + top-3}}  & $\mathbf{0.15 \pm 0.01}$ & \textbf{(9.3, 9.7)} & \textbf{330}  \\
\bottomrule
\end{tabular}
\caption{Results for finetuned open source models with our TinyTolkien dataset. }
\label{tab:ft_results}
\end{table*}

\subsection{Finetuning approaches}

In contrast to prompt-based strategies, we can also directly finetune open source LLMs for the PCT. Finetuned LLMs can be more efficient in terms of token usage cost and have the potential to match the performance of proprietary models. The major limitation of this approach is that it requires a gold-standard dataset of tuples $\{(p_i, t_i, x_i)\}_{i=1}^n$, where $p_i \in \Sigma^*$ is a prompt, $t_i \in \{1, 2, \ldots, 6\}$ is a target proficiency, and $x_i \in \Sigma^*$ is a gold standard response to the prompt at proficiency level $t_i$. 



Given this kind of dataset, we can finetune a model using the standard causal language modelling objective. Following prior work on controllable generation \cite{keskarCTRL2019,stowe-etal-2022-controlled}, we append the target proficiency level as a control token after the prompt. At test time, this token can be chosen to generate at any target proficiency level.

\subsection{Proximal Policy Optimisation (PPO) for Proficiency Alignment}

Finetuning with control tokens can improve the controllability of an LLM. However, the generated responses might not be well-aligned with the target proficiency. Recent work  \cite{rlhf22} has shown promising results in using reinforcement learning algorithms like Proximal Policy Optimisation (PPO) \cite{schulman2017proximal} to further align the outputs of a model with an objective function. In the case of the PCT, we can use the negative of the $\CS$ of a given generation as a reward in the PPO algorithm to incentivise generations that closer match the target level.

\subsection{Boosting Models with top-$k$ Sampling}
All LLMs use a stochastic sampling strategy to generate text. This means, for a given prompt and target level, a PCT model could generate responses with varying degrees of $\CS$. This suggests an easy method to reduce the $\CS$ of \textit{any} PCT model: {sample $k$ random responses for a given prompt and target level, and return the one with the lowest $\CS$}. A similar technique was used in \citet{ribeiro-etal-2023-generating}.


The top-$k$ algorithm provably reduces the $\CS$ of a PCT model, but incurs a higher cost since it requires several generation requests for one prompt. In \Cref{sec:boosting}, we show how this simple approach can boost an acceptable but cost-effective model into an extremely powerful one.

\section{Experimental Setup}

To experiment with the different proficiency control strategies, we run an experiment using the TinyStories dataset \cite{eldan2023tinystories}, which is a collection of English short-stories that also includes a plot summary for each story (CDLA-Sharing-1 license). Using this data, we construct the following task: a model is given the plot summary of a story as well as a uniformly random target CEFR level from 1 to 6. The model is then asked to generate a short story (around 3-5 paragraphs) which adheres to the given plot and also sits at the target level. We select a subset of $50$ random story plots from the TinyStories dataset to evaluate on. See \Cref{appx:exp_details} for all training details.\footnote{We also release our code, datasets, and finetuned models in a public repository.}

\subsection{Evaluation metrics}

According to our PCT framework, we need to measure the average $\CS$, $\QS$, and $\Cost$ of each proficiency control strategy. 
We can measure the $\CS$ of the generated story directly using our automatic scoring function.

To measure $\QS$, we use the same evaluation framework as the TinyStories paper. For each story plot and generated story, we ask GPT-4 to grade the text in terms of both language fluency and consistency with the given plot. Following \citet{chiang-lee-2023-large}, we expect these to have high correlation with human judgements, but we also validate this with a human study (\Cref{sec:human-eval}). Both these quantities are scored on a scale from 1-10 and reported as a tuple of $(\mathrm{Fluency}, \mathrm{Consistency})$. 

Lastly, we measure $\Cost$ of a strategy using an estimate of floating-point operations (FLOPs), which is a measure of how much compute is used to generate a story at a target level for a given prompt. The FLOP estimate is a function of tokens generated, and number of parameters in the model, under the assumption that all parameters are used to generate each token. For open-source models, we compute FLOPs using the published number of parameters. For GPT-4, the details are hidden and we have no recourse but speculation. GPT-3 has 175B parameters \cite{brown2020language}, and we may reasonably assume that GPT-4 is larger. 
Thus, when comparing relative costs between GPT-4 and the 7B parameter models, a factor of $175/7 = 25$x is the lower bound.



\section{Results: Prompt-based Approaches}\label{sec:prompt}




In \Cref{tab:prompt_results}, we evaluate all the different combination of prompting strategies from \Cref{ssec:prompt_strategies}, each labelled by a letter, on OpenAI's GPT-4 \cite{openai2024gpt4}, LLaMa-2-7b-chat \cite{touvron2023llama}, and Mistral-7b-instruct \cite{jiang2023mistral}. For each strategy and model combination, we report the $\CS$ with standard error, the $\QS$ represented as a tuple of the $(\mathrm{Fluency}, \mathrm{Consistency})$ scores out of 10, and number of tokens needed for each strategy. We do not report standard errors for the $\QS$ because they are all effectively $0$.
We observe several interesting findings:

\paragraph{(1) Quality and scale of the LLM matters} We see a stark performance gap between GPT-4 and the open source models at controlling CEFR proficiency. Even using the most complex prompting strategies, the performance of the open source models is poor compared to the most basic prompt for GPT-4. This suggests that the quality and scale of the underlying model matters.

\paragraph{(2) More details improve proficiency control} For GPT-4, we see a decrease in the $\CS$ as we provide more detail about CEFR levels in the prompt. For example, adding a description of the target CEFR level or including few-shot examples reduces the $\CS$ significantly. 


\paragraph{(3)  Quality is consistently high} Looking at the fluency and consistency of the generated stories, we observe high scores across all models and all strategies. This is promising evidence that all these models are good at the story generation task, albeit with varying proficiency control capabilities.

\section{Distilling GPT-4 for Open Source}\label{sec:ft}

The high $\QS$ but low $\CS$ of the open source models suggests that they are quite capable at story generation, but lack the ability to be steered through prompting. A promising path forward is to directly finetune these models for controllable CEFR generation. Following a similar idea to TinyStories, we investigate whether GPT4's effectiveness at the PCT can be leveraged to improve the open source models.

\subsection{The TinyTolkien Dataset}
\label{sec:tinytolkien}
To make progress on this front, we use the GPT4(b) strategy (\Cref{tab:prompt_results}) to generate reference stories to given plots from TinyStories at different CEFR proficiency levels. 

\begin{figure}
    \centering
    \includegraphics[width=\linewidth]{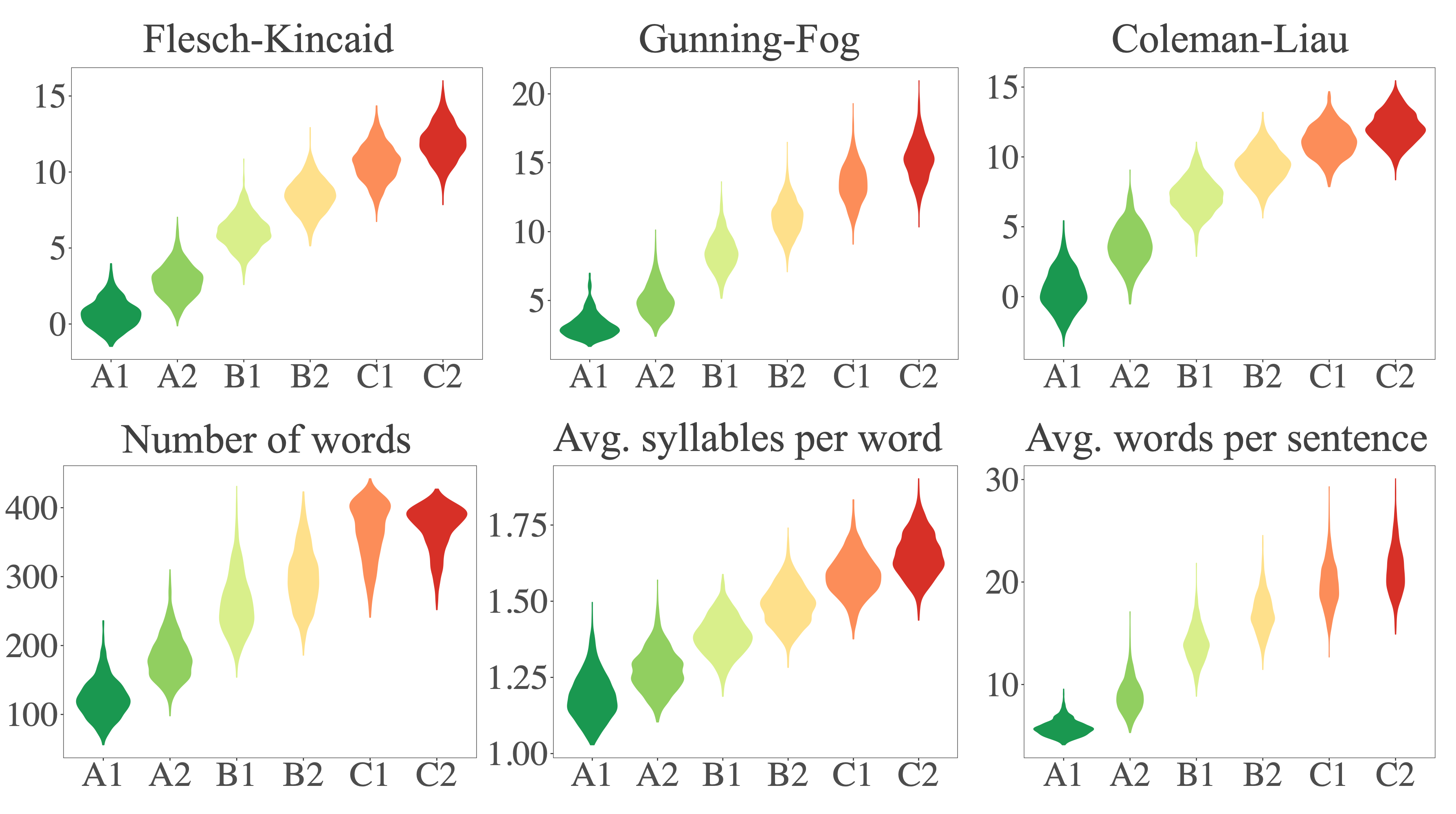}
    \caption{Distribution of different readability metrics for each CEFR level in the generated TinyTolkien data.}
    \label{fig:tinytolkien_metrics}
\end{figure}

Specifically, we sample a random subset of $1000$ story plots, and for each one, we select two random target CEFR levels to generate, resulting in a total of $2000$ data points.
We call this data the TinyTolkien dataset and use it for the rest of our paper. Some readability metrics for text at each target level are included in \Cref{fig:tinytolkien_metrics} and examples of the data can be found in \Cref{appx:tinytolkien}.

\subsection{Finetuning}

\Cref{tab:ft_results} shows the results for LLama2-7b and Mistral-7b after finetuning on the TinyTolkien dataset. We observe almost a $50\%$ reduction in $\CS$ of the finetuned models compared to their original versions with prompting while still retaining their high $\QS$.

\subsection{Proximal Policy Optimisation (PPO)}

Although the finetuned models show improved performance, they still lag behind the GPT-4(b) strategy. Our investigations reveal that the finetuned models exhibit a clear degree of proficiency control, but the outputs are misaligned with respect to the prediction given by our CEFR scoring function.

To further align the model output proficiency, we run Proximal Policy Optimisation (PPO) by using the negative of the $\CS$ as a reward function. We find PPO to greatly improve the $\CS$ performance of both open source models, resulting in a \textit{further} $50\%$ decrease in $\CS$ of LLama2-7b without affecting quality (\Cref{tab:ft_results}). In particular, we are able to bring the LLaMa2-7b model to match the performance of the GPT4(b) strategy. \textbf{We call this final model \Calm{} for CEFR-aligned Language Model.}

Despite these improvements, it is important to note that the PPO training is highly unstable. Training the model for too long causes the outputs to degenerate into repeating sequences or nonsensical bytes. We share the training details of our PPO and finetuning in a public code repository.

\section{Boosting Models Using top-$k$ Sampling}\label{sec:boosting}

\begin{figure}
    \centering
    \includegraphics[width=0.99\linewidth]{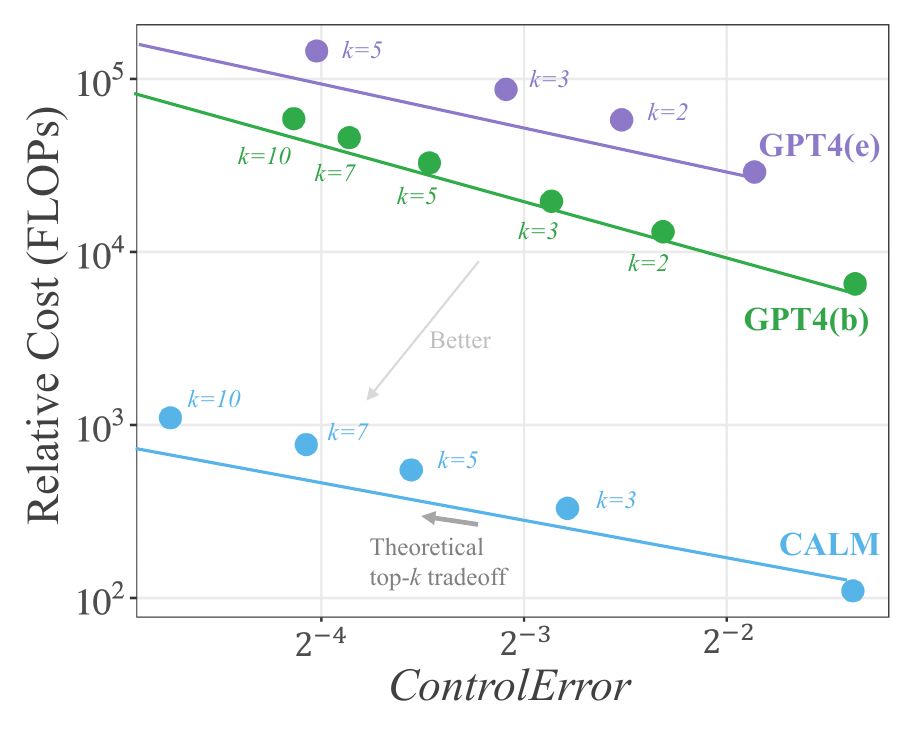}
    \caption{Tradeoff between relative cost (in FLOPs) and $\CS$ for different strategies. Each base point represents a different strategy, and additional points per colour show results for top-$k$ sampling with that strategy. Increasing $k$ reduces the error of any strategy by paying a higher cost. The solid lines represent the theoretical trade-off (estimated using bootstrap sampling) in cost vs $\CS$ as $k$ is increased for each strategy.}
    \label{fig:tradeoff}
\end{figure}

All PCT models discussed above naturally exhibit randomness in their generations. This suggests an easy way to reduce the $\CS$ of any such model: sample $k$ independent generations for a prompt and choose the one with the lowest $\CS$. 
Although this strategy is extremely simple, it leads to a powerful new capability: for any PCT model, we can pay a higher cost (by increasing $k$) and in turn reduce our $\CS$. 

The existence of this $\Cost$ vs $\CS$ trade-off suggests a need for an optimality analysis between the different techniques when combined with top-$k$. To answer this, we construct a cost/error trade-off plot for each strategy.
\Cref{fig:tradeoff} shows this trade-off plot for some of our key PCT strategies, as well as how this changes when combined with top-$k$ for $k = 2, 3, 5, 7$ and $10$. We also compute a theoretical trade-off curve (solid lines on plot) for how the error/cost of each prompting strategy would change when combined with top-$k$ sampling, for increasing values of $k$. This is estimated using bootstrap sampling.

Looking at the \Cref{fig:tradeoff}, we see a striking result. Our \Calm{} model strictly dominates all the GPT-4 prompt-based strategies in terms of $\CS$ and $\Cost$. In other words, it is always cheaper to use \Calm{} + top-$k$ sampling to attain whatever $\CS$ is desired.

\begin{table}[t]
    \small
  \centering
  \begin{tabular}{p{0.1\textwidth} p{0.12\textwidth}  p{0.11\textwidth}}
  \toprule
  \textbf{Model}  & \textbf{Consistency Rating} {(1-5}) &\textbf{Language Rating} {(1-5)}  \\
  \midrule
    {GPT4(b)}  & $4.8 \pm 0.1$ & $4.6 \pm 0.1$ \\
    {\Calm{}}  & $4.7 \pm 0.1$ & $4.3 \pm 0.1$ \\
\bottomrule 
\end{tabular}
\caption{Human evaluation of the quality of generated stories of GPT4(b) and \Calm. Both models are rated highly in terms of consistency with the plot and quality of language used.}
\label{table:human-ratings}
\end{table}

\section{Human Evaluations}
\label{sec:human-eval}
In order to further validate our results from the automatic CEFR scorer and the GPT-4 based evaluation of quality, we ran a small human evaluation. We recruited 13 volunteer participants from peers and colleagues to do a blind evaluation. We gave the volunteers two tasks. 

\subsection{Quality of Generated Stories}

In the first task, participants were asked to give absolute ratings of generated stories, rating both \textbf{Consistency} and \textbf{Language Score} on a scale of 1 to 5. The former measures how consistent the generated story is with the plot summary in the prompt, and the latter measures how fluent the story is in terms of correct use of English grammar and sentences. The instructions we gave to raters are included in \Cref{sec:eval-instructions}

We evaluated generations from two PCT models: GPT4(b), and \Calm{}. The results can be seen in \Cref{table:human-ratings}.
We see that evaluators rated both PCT models highly in terms of consistency and use of language. 
In terms of evaluator reliability, the expected squared distance in ratings between two random evaluators was about $0.2$ for the consistency score and about $0.87$ for the language score.

\begin{figure}
    \centering
    \includegraphics[width=\linewidth]{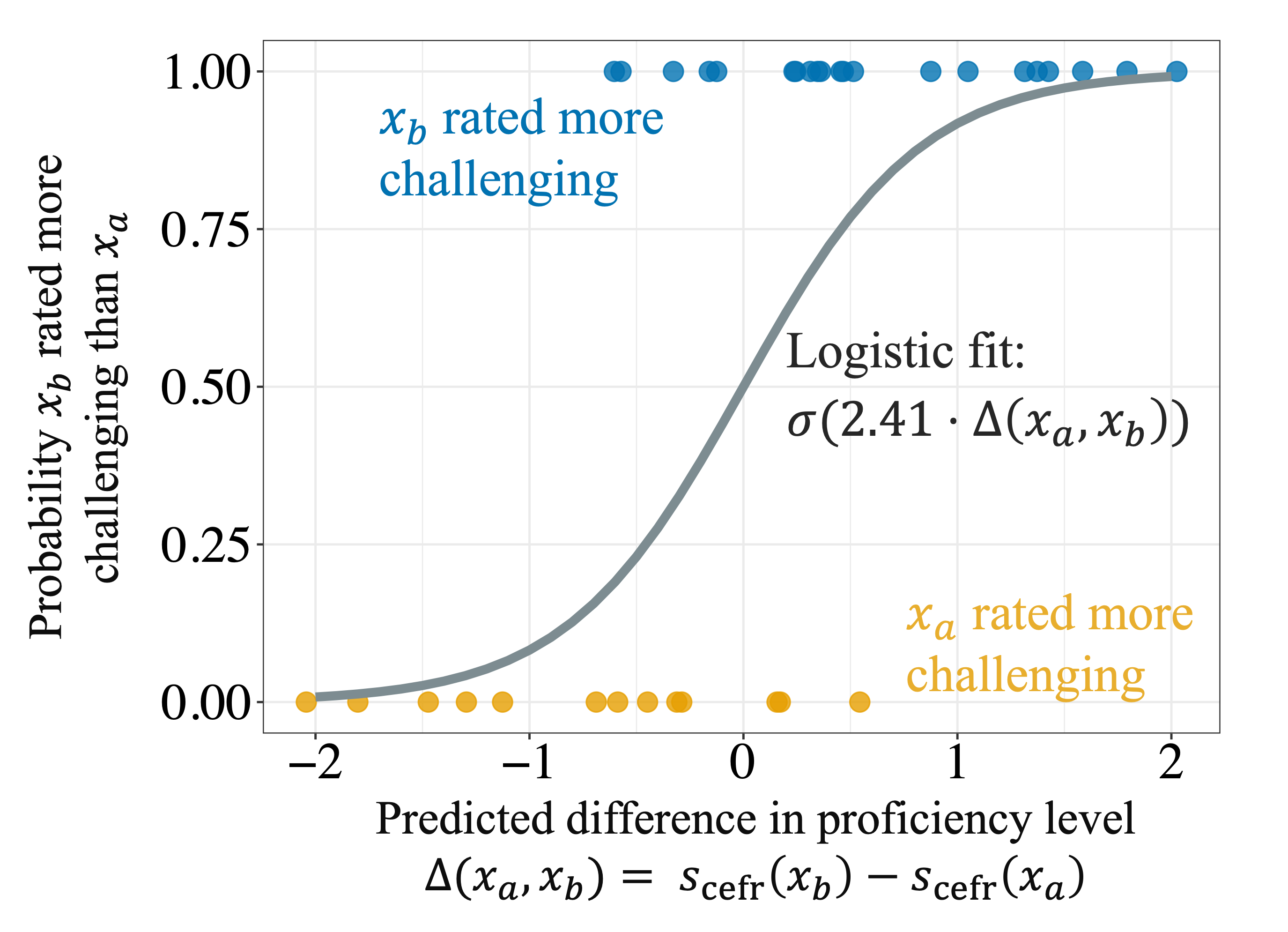}
    \caption{\textbf{Predicted CEFR scores correspond to human perception of difficulty}. As the difference in predicted proficiency scores between story A and story B increases, humans are better able to identify the more challenging story. The yellow dots (y = 0) correspond to instances where the evaluator rated story A as more challenging and the blue dots (y = 1) correspond to when they rated story B as more challenging. }
    \label{fig:human-prof}
\end{figure}

\subsection{Automatic CEFR Scorer}
We also looked at how well our CEFR scoring function matched with human perceptions of proficiency levels. In the second evaluation task, participants were shown two stories, and asked which of the two was more challenging in terms of English proficiency level. Behind the scenes, the stories were generated using \Calm{} at two random target CEFR levels. We looked at how well participants could identify the more challenging story as a function of how much higher our CEFR scorer rated one over the other.

\Cref{fig:human-prof} shows a summary of this evaluation. The yellow dots (y = 0) correspond to instances where the evaluator rated Story A as more challenging and the blue dots (y = 1) correspond to when they rated Story B as more challenging. The $x$-axis plots the difference in predicted proficiency scores between Story B and Story A, as measured by our automatic scorer.

We see a clear trend in the human evaluators' ability to distinguish between proficiency levels. As the predicted difference in proficiency levels gets larger,  humans are better able to distinguish between the two stories. In fact, we find a clear fit with a logistic regression for the probability that an evaluator chooses Story B as more challenging, as a function of this difference. This suggests that our automated scoring function exhibits clear predictive power over human perception of difficulty. The graph also suggests that a $\CS$ of about $0.25$ is about the most granularity needed before generations are imperceptible to humans.

\begin{table*}
\centering
\small
    \begin{tabular}{c | p{0.85\linewidth}}
    \toprule
        \textit{Prompt} & \textit{Tom borrows Lily's favorite toy truck but gets distracted by a real truck outside and runs out, making Lily sad. Tom gets scared when the truck driver yells at him and runs back to Lily's room.} \\
            \midrule
            
    \textbf{Target (Model)} &  \textbf{Output} \\

    \midrule
    A1 (GPT4) & Tom takes Lily's toy truck. Lily loves this truck. But Tom sees a big, real truck outside. He runs out to see it. Lily is sad...  
    \\
    \\
    A1 (\Calm{}) & Tom likes a toy truck. One day, he sees Lily's toy truck. He wants to play with it. So, he goes to Lily's room. But, he sees a real truck outside. The truck is big and loud... 
\\
    \midrule
    B1 (GPT4) & One sunny day, Tom was at Lily's house. Lily had a toy truck that was her favorite, and she allowed Tom to play with it. Tom was enjoying playing with the toy truck when a big, noisy truck drove by outside... 
    \\
    \\
     B1 (\Calm{}) & Tom was playing with Lily's favorite toy truck when he saw a real, big truck outside the window. It was bright red and loud, and it caught his attention. He forgot about the toy truck and ran outside to see the real truck... 
  \\
  \midrule
    C1 (GPT4) & Tom had always been captivated by the enchanting world of toys. His friend Lily owned a striking vermilion toy truck, a miniature replica of a Mack Titan, that had become the apple of Tom's eye. One day, while engrossed in play, Tom managed to convince Lily to lend him her favorite toy truck...
    \\
    \\
    C1 (\Calm{}) & In the cozy corner of Lily's bedroom, where the dusty sunlight peeked in through the window, Tom had found a cherished toy truck. The toy truck, a replica of a huge cargo truck, was painted vibrant red and yellow with a black grille, and it was Lily's favorite... 
\\
    \bottomrule
    \end{tabular}
    \caption{Outputs of the \textbf{GPT4(b) + top-3} and \textbf{\Calm{}} strategies at different target levels.}
\end{table*}

\section{Related Work}
\label{sec:rw}


\subsection{Language Proficiency Standards}
In addition to the Common European Framework of Reference (CEFR) \cite{councilofeurope2001}, other language standards include Interagency Language Roundtable (ILR) (used by \cite{salesky-shen-2014-exploiting}), and ACTFL, used primarily in the United States.\footnote{https://www.languagetesting.com/cefr-scale}
We choose to use the CEFR because of its wide adoption in language learning and language proficiency testing \cite{settles2020,mccarthy-etal-2021-jump}.

When discussing CEFR, we make a distinction between the different texts that might need classification (as seen in \citep{pv18investigating})

\begin{enumerate}
    \item L1 text aimed at \textbf{natives}, made by teachers, such as books for small children
    \item L2 text aimed at \textbf{learners}, made by teachers, including most language learning materials
    \item L2 text produced by learners, such as language exam question responses
\end{enumerate}

Since we are focused on language learning, we primarily target the second type, with the LLM as a stand-in for the ``teacher." 

Several datasets with CEFR labels exist -- automatically aligned English/Simple English Wikipedia \cite{wilkens-etal-2018-sw4all}, automatically and human-tagged learner texts \cite{tack-etal-2017-human}, and others \cite{xia-etal-2016-text, montgomerieCEFRdata,Breuker2023}, almost always in English.

\paragraph{Automatic proficiency evaluation of text} 
Automatic language proficiency evaluation is a well-studied question, and a thorough overview can be found in \citet{pv18investigating}. Several prior works find a similar common set of features that are highly predictive of proficiency level, including Text-to-Token ratio \cite{pv18investigating}, morphological, information-theoretic, and language modeling features \cite{salesky-shen-2014-exploiting, xia-etal-2016-text}, part of speech and dependency parse \cite{vr18univclass}, word frequency and expert knowledge \cite{pf20combining}. Recent works also explore ensemble methods \cite{tack-etal-2017-human} and deep learning \cite{deutsch-etal-2020-linguistic,kerz-etal-2021-automated}.






\paragraph{Simplification and Readability}
Text simplification and readability assessment, while not directly related to the PCT, have many similarities to the task. In particular, recent works have addressed text simplification with a particular target in mind \cite{scarton-specia-2018-learning, kew-ebling-2022-target, agrawal-carpuat-2023-controlling}  \cite{agrawal-carpuat-2019-controlling} adopt a multi-task machine translation and simplification framework to do ``complexity controlled machine translation.'' 

While most work on readability assessment is in English, some works have expanded to other languages including Russian \cite{reynolds-2016-insights}, Bangla \cite{islam-rahman-2014-readability}, and Philippine languages \cite{imperial-kochmar-2023-automatic,imperial-kochmar-2023-basahacorpus}

Concurrent to our work, \citet{ribeiro-etal-2023-generating} explore summarization with fine-grained control over readability. As in our work, they find that prompting can be successful, but additional RL-based methods, as well as lookahead decoding improve the results further. They also propose a top-$k$ sampling approach similar to ours for GPT3.5.

\paragraph{Controllable generation} As generative language models have become more popular, interest in controllable text generation has increased. In a survey on Controllable Text Generation, \citet{ctg-survey-2022} list several common applications, including attribute-based generation (e.g. politeness \cite{sennrich-etal-2016-controlling}), storytelling \cite{Prabhumoye2019MyWO}, and format control \cite{li-etal-2020-rigid}. Our work falls under attribute-based generation, with the attribute being CEFR level.
CTRL \citep{keskarCTRL2019} used \textit{control codes} prepended to each training sequence to direct the model in a certain direction, evaluating on topics such as \texttt{Wikipedia} and \texttt{Legal}, but not language proficiency.


In one of the earlier studies of CEFR-controlled generation, \citet{stowe-etal-2022-controlled} explore controlled generation for Language Learning Applications using a concept2seq framework, with control features such as CEFR and Semantic Role Labels, and the encoder-decoder framework, BART, as their model. They limit the CEFR task to the extremes, and only generate in A1 or C2, showing good success in differentiation. We build on their work by broadening the CEFR generation task to all labels, and by using a new era of prompt-based LLMs. 


Concurrently to our work, \citet{imperial2023Flesch} use a variety of open-source and proprietary models to explore prompting methods on two sub-tasks: open-ended story completion, and narrative simplification. As in our work, they find that LLMs with no specific proficiency instructions produce high-fluency level text, but that the more information given in the prompt, the better the results. Our work goes beyond theirs in experimenting with a broader scope of target proficiency levels, and also on both simplifying \textit{and} complexifying text. We also further explore finetuning as a way to empower smaller, open source models.

\section{Conclusion}
We present a new challenge for controlling the proficiency level of LLM generated content: a highly practical and important task for practitioners in the domain of education and/or language learning. We demonstrate effective strategies for generating at a desired proficiency level, using both proprietary models such as GPT4 and open source techniques. Through a careful cost analysis, we show that our \Calm{} model is dominant in terms of cost and performance, and generates content rated by humans to be of high quality. We release this model as well as a synthetic toy dataset called TinyTolkien for future use in proficiency control research.

\section{Limitations}

\subsection{CEFR Ambiguity}

One challenge for any research in this area is the inherent ambiguity in the CEFR scale. While it is useful in broad strokes, and while there is very little confusion between, say, A1 and C2, for many texts (especially short texts), there is no consistent, coherent process that places them firmly in one of two adjacent CEFR levels.

This ambiguity is reflected in our automatic proficiency scoring function, and consequently in the evaluation of the main prompting strategies of this paper. However, this is a function of the task, not of the solution. This problem will remain until an unambiguous proficiency framework is created.

\subsection{Difficulty vs Fluency}
A related challenge to the ambiguity of CEFR is differentiating between the role of ``fluency'' and ``difficulty'' in the different levels. While one way of interpreting C2 text is in terms of the complexity of the content, it could also be used as a measure of the fluency of the writing. In this sense, a masterful C2 level text could be simple to read, but successfully capture nuances and intricate ideas. On the other hand, the C2 text according to most automated scoring functions is often unnaturally complex and relies on long sentence constructions and obscure words. Reasoning about what proficiency truly means is an important pedagogical and philosophical question for further work in this area.

\subsection{Evaluation with closed models}
A portion of our results come from outputs of closed systems, over which we have no control. As models are updated and deprecated, these exact results may prove hard to reproduce. Given the importance of these models in the field and the world, we thought it important to evaluate them despite these risks.

\subsection{Generalising to other languages}

The majority of this work was focused on proficiency control \textit{in the context of English}. However, we hope the methods here easily generalise to other languages. There are certain challenges. Firstly, the ability to train an automatic CEFR scorer requires a labelled dataset of CEFR texts. These are more readily available in certain popular languages like English. Extending this work to the low-resource language setting is an exciting future direction.

\subsection{Biases in AI-generated data}
Both the original TinyStories dataset \cite{eldan2023tinystories} that we experiment with and our TinyTolkien dataset are AI generated. Data generated from LLMs has the potential to exhibit and promote biases \cite{Fang2024}. For example, we observe that the stories in TinyStories tend to use predominantly western names such as Jack and Mary that are common in classical children's stories. The extension of this data with TinyTolkien exhibits a similar pattern. While we use this data as a testing ground for our ideas, care should be taken to deploy models for content generation in the real world.


\bibliography{anthology,custom}
\bibliographystyle{acl_natbib}

\appendix
\onecolumn

\section{CEFR Level Descriptions}

\begin{table*}[h]
\centering
\begin{tabular}{p{0.05\linewidth} | p{0.95\linewidth}}
\toprule
\textbf{Level} & \textbf{Description} \\
\midrule
A1 & I can understand familiar names, words and very simple sentences, for example on notices and posters or in catalogues.  \\ 
A2 & I can read very short, simple texts. I can find specific, predictable information in simple everyday material such as advertisements, prospectuses, menus and timetables and I can understand short simple personal letters.  \\ 
B1 & I can understand texts that consist mainly of high frequency everyday or job-related language. I can understand the description of events, feelings and wishes in personal letters. I can read articles and reports concerned with contemporary problems in which the writers adopt particular attitudes or viewpoints.  \\ 
B2 & I can understand contemporary literary prose. I can understand long and complex factual and literary texts, appreciating distinctions of style.  \\ 
C1 & I can understand specialised articles and longer technical instructions, even when they do not relate to my field. \\ 
C2 & I can read with ease virtually all forms of the written language, including abstract, structurally or linguistically complex texts such as manuals, specialised articles and literary works. \\ 
\bottomrule
\end{tabular}
\caption{Official ``can-do`` descriptors for reading-based understanding at different CEFR levels (Council of Europe)}
\label{table:cefr_levels}
\end{table*}



\section{Experimental details}\label{appx:exp_details}
In this section, we provide all the experimental details, prompts, hyperparameters, sampling parameters, and training details used in our results.

\subsection{Automatic CEFR Scorer}\label{appx:scorer_details}

\paragraph{Datasets.} We gather three different datasets of CEFR levelled English texts. The first is the EDIA/European Language Grid dataset \cite{Breuker2023}, which consists of around 1200 texts from various sources, labelled on CEFR readability level. A few texts are labelled between levels, which we round down. The second dataset is the CambridgeExams dataset \cite{xia-etal-2016-text}, which is ``composed of reading passages from the five main suite Cambridge English Exams ... targeted at learners at A2–C2''. This dataset consists of 331 texts spanning levels A2 through C2, with roughly 60 documents per level. Lastly, we look at a Kaggle dataset gathered from free online resources such as The British Council, ESLFast, and the cnn-dailymail \cite{montgomerieCEFRdata}. This is the largest dataset, with around 1500 texts, but some of the entries are labelled using a paid automated labelling service.

For our experiments, we use a scorer based on a combination of the EDIA and Kaggle dataset. We unforunately only learned about the CambridgeExams dataset after running our experiments, but otherwise have incorporated it as well. See \Cref{appx:scorer_disc} on a discussion of the robustness of our results to different scoring functions.

\paragraph{Features.} We featurised each text using a set of common linguistic features. The features can be categorised into three main groups, focusing on word frequency, syntactic complexity, and part-of-speech (POS) distribution. It is important that our features don't depend on the length of the text, otherwise algorithms like PPO would exploit this by generating shorter or longer sentences to attain a target level.

\begin{enumerate}
    \item Word Frequency Bins: We compute rank bins e.g. rank\_0\_250, rank\_250\_500, etc. that represent the distribution of words across various frequency bins. Each bin encompasses a range of word ranks based on their frequency in the Oxford English Corpus, with higher ranks indicating less frequent words.
    \item Syntactic Complexity Measures: We compute measures such as Average Sentence Length, Average Maximum Parse Tree Depth, Average Maximum Children, Average Number of Unique Dependencies. 
    \item Part-of-Speech Tagging averages: These features represent the average distribution of various POS tags across all sentences.
\end{enumerate}

We elected to use a straightforward set of features for simplicity and to avoid PPO exploiting idiosyncrasies in a more complex scoring function. Nevertheless, we believe our results would generalise with any reasonable scoring function.

\subsection{Prompting strategies}\label{appx:prompts}

We share the prompt used for each of the prompting strategies. We use A1 as the example target level

\textbf{{(a): Base}}
\begin{lstlisting}
====== System ======
You are a writer that generates a story according to a given plot summary.
====== User ======
Generate according to the prompt below but make sure that the generated text is at the A1 level of English proficiency.

Write a short story (3-5 paragraphs) with the following plot. Output the story only and no other text!

Plot: {Story plot}
\end{lstlisting}

\textbf{(b): Target description}
\begin{lstlisting}
====== System ======
You are a writer that generates a story according to a given plot summary.
====== User ======
Generate according to the prompt below but make sure that the generated text is at the A1 level of English proficiency.

As a reminder, A1 proficiency is described as:

## A1 (Beginner)
The writing uses familiar names, words and very simple sentences, for example as seen on notices and posters or in catalogues.
- Includes the top most frequent 1,000 commonly spoken words in the language
- Includes many words and phrases that fall under common early language learning topics (e.g. common greeting, travel, dining, shopping, etc)
- Includes all proper nouns (country names, person names, etc)
- Includes all cognates shared with English
- Includes all words that look similar to English words that share a similar meaning


---------------------------------------------------
Prompt:

Write a short story (3-5 paragraphs) with the following plot. Output the story only and no other text!

Plot: {story plot}
\end{lstlisting}

\textbf{(c): Target description + Target example}
\begin{lstlisting}
====== System ======
You are a writer that generates a story according to a given plot summary.
====== User ======
Generate according to the prompt below but make sure that the generated text is at the A1 level of English proficiency.

As a reminder, A1 proficiency is described as:

## A1 (Beginner)
The writing uses familiar names, words and very simple sentences, for example as seen on notices and posters or in catalogues.
- Includes the top most frequent 1,000 commonly spoken words in the language
- Includes many words and phrases that fall under common early language learning topics (e.g. common greeting, travel, dining, shopping, etc)
- Includes all proper nouns (country names, person names, etc)
- Includes all cognates shared with English
- Includes all words that look similar to English words that share a similar meaning

Example 1: {A1 example}

---------------------------------------------------
Prompt:


Write a short story (3-5 paragraphs) with the following plot. Output the story only and no other text!

Plot: {Story plot}
\end{lstlisting}

\textbf{(d): All levels description}
\begin{lstlisting}
====== System ======


You are a large language model that can generate content at a certain proficiency level suitable for English language learners.
Your goal is to output content and text at the proficiency level specified in the prompt.

The descriptions of the proficiency levels are given as follows:

## A1 (Beginner)
The writing uses familiar names, words and very simple sentences, for example as seen on notices and posters or in catalogues.
- Includes the top most frequent 1,000 commonly spoken words in the language
- Includes many words and phrases that fall under common early language learning topics (e.g. common greeting, travel, dining, shopping, etc)
- Includes all proper nouns (country names, person names, etc)
- Includes all cognates shared with English
- Includes all words that look similar to English words that share a similar meaning


## A2 (Elementary)
The writing involves short, simple texts with specific, predictable information. Examples include simple everyday material such as advertisements, prospectuses, menus and timetables or short simple personal letters.
- Includes the top most frequent 1,000-2,000 commonly spoken words in the language


## B1 (Intermediate)
Texts that consist mainly of high frequency everyday or job-related language. These involve descriptions of events, feelings and wishes in personal letters.
- Includes the top 2,000-5,000 commonly spoken words in the language
- Includes several rarer verb tenses (e.g. conditional, subjunctive, etc)
- Includes some relatively common idiomatic phrases


## B2 (Upper Intermediate)
Writing as seen in articles and reports concerned with contemporary problems in which the writers adopt particular attitudes or viewpoints. Also includes contemporary literary prose.
- Includes the top 5,000-10,0000 commonly spoken words in the language


## C1 (Proficient)
Writing can include long and complex factual and literary texts, with distinctions of style. Examples include specialised articles and longer technical instructions, even when they do not relate to a well-known field.
- Includes the top 10,0000-20,0000 commonly spoken words in the language


## C2 (Advanced Proficient)
Includes virtually all forms of the written language, including abstract, structurally or linguistically complex texts such as manuals, specialised articles and literary works.
- Includes esoteric technical language


--------------------------------------------------
You are a writer that generates a story according to a given plot summary.
====== User ======
Generate according to the prompt below but make sure that the generated text is at the A1 level of English proficiency.

Write a short story (3-5 paragraphs) with the following plot. Output the story only and no other text!

Plot: {story plot}
\end{lstlisting}

\textbf{(e): All levels description + target example}
\begin{lstlisting}
====== System ======


You are a large language model that can generate content at a certain proficiency level suitable for English language learners.
Your goal is to output content and text at the proficiency level specified in the prompt.

The descriptions of the proficiency levels are given as follows:

## A1 (Beginner)
The writing uses familiar names, words and very simple sentences, for example as seen on notices and posters or in catalogues.
- Includes the top most frequent 1,000 commonly spoken words in the language
- Includes many words and phrases that fall under common early language learning topics (e.g. common greeting, travel, dining, shopping, etc)
- Includes all proper nouns (country names, person names, etc)
- Includes all cognates shared with English
- Includes all words that look similar to English words that share a similar meaning


## A2 (Elementary)
The writing involves short, simple texts with specific, predictable information. Examples include simple everyday material such as advertisements, prospectuses, menus and timetables or short simple personal letters.
- Includes the top most frequent 1,000-2,000 commonly spoken words in the language


## B1 (Intermediate)
Texts that consist mainly of high frequency everyday or job-related language. These involve descriptions of events, feelings and wishes in personal letters.
- Includes the top 2,000-5,000 commonly spoken words in the language
- Includes several rarer verb tenses (e.g. conditional, subjunctive, etc)
- Includes some relatively common idiomatic phrases


## B2 (Upper Intermediate)
Writing as seen in articles and reports concerned with contemporary problems in which the writers adopt particular attitudes or viewpoints. Also includes contemporary literary prose.
- Includes the top 5,000-10,0000 commonly spoken words in the language


## C1 (Proficient)
Writing can include long and complex factual and literary texts, with distinctions of style. Examples include specialised articles and longer technical instructions, even when they do not relate to a well-known field.
- Includes the top 10,0000-20,0000 commonly spoken words in the language


## C2 (Advanced Proficient)
Includes virtually all forms of the written language, including abstract, structurally or linguistically complex texts such as manuals, specialised articles and literary works.
- Includes esoteric technical language


--------------------------------------------------
You are a writer that generates a story according to a given plot summary.
====== User ======
Generate according to the prompt below but make sure that the generated text is at the A1 level of English proficiency.

As a reminder, A1 proficiency is described as:

## A1 (Beginner)
The writing uses familiar names, words and very simple sentences, for example as seen on notices and posters or in catalogues.
- Includes the top most frequent 1,000 commonly spoken words in the language
- Includes many words and phrases that fall under common early language learning topics (e.g. common greeting, travel, dining, shopping, etc)
- Includes all proper nouns (country names, person names, etc)
- Includes all cognates shared with English
- Includes all words that look similar to English words that share a similar meaning

Example 1: {A1 example}

---------------------------------------------------
Prompt:


Write a short story (3-5 paragraphs) with the following plot. Output the story only and no other text!

Plot: {story plot}   
\end{lstlisting}

\textbf{(f): All levels description + all levels example}
\begin{lstlisting}
====== System ======


You are a large language model that can generate content at a certain proficiency level suitable for English language learners.
Your goal is to output content and text at the proficiency level specified in the prompt.

The descriptions of the proficiency levels are given as follows:

## A1 (Beginner)
The writing uses familiar names, words and very simple sentences, for example as seen on notices and posters or in catalogues.
- Includes the top most frequent 1,000 commonly spoken words in the language
- Includes many words and phrases that fall under common early language learning topics (e.g. common greeting, travel, dining, shopping, etc)
- Includes all proper nouns (country names, person names, etc)
- Includes all cognates shared with English
- Includes all words that look similar to English words that share a similar meaning

Example 1: {A1 example}

## A2 (Elementary)
The writing involves short, simple texts with specific, predictable information. Examples include simple everyday material such as advertisements, prospectuses, menus and timetables or short simple personal letters.
- Includes the top most frequent 1,000-2,000 commonly spoken words in the language

Example 1: {A2 example}

## B1 (Intermediate)
Texts that consist mainly of high frequency everyday or job-related language. These involve descriptions of events, feelings and wishes in personal letters.
- Includes the top 2,000-5,000 commonly spoken words in the language
- Includes several rarer verb tenses (e.g. conditional, subjunctive, etc)
- Includes some relatively common idiomatic phrases

Example 1: {B1 example}

## B2 (Upper Intermediate)
Writing as seen in articles and reports concerned with contemporary problems in which the writers adopt particular attitudes or viewpoints. Also includes contemporary literary prose.
- Includes the top 5,000-10,0000 commonly spoken words in the language

Example 1: {B2 example}

## C1 (Proficient)
Writing can include long and complex factual and literary texts, with distinctions of style. Examples include specialised articles and longer technical instructions, even when they do not relate to a well-known field.
- Includes the top 10,0000-20,0000 commonly spoken words in the language

Example 1: {C1 example}

## C2 (Advanced Proficient)
Includes virtually all forms of the written language, including abstract, structurally or linguistically complex texts such as manuals, specialised articles and literary works.
- Includes esoteric technical language

Example 1: {C2 example}

--------------------------------------------------
You are a writer that generates a story according to a given plot summary.
====== User ======
Generate according to the prompt below but make sure that the generated text is at the A1 level of English proficiency.

As a reminder, A1 proficiency is described as:

## A1 (Beginner)
The writing uses familiar names, words and very simple sentences, for example as seen on notices and posters or in catalogues.
- Includes the top most frequent 1,000 commonly spoken words in the language
- Includes many words and phrases that fall under common early language learning topics (e.g. common greeting, travel, dining, shopping, etc)
- Includes all proper nouns (country names, person names, etc)
- Includes all cognates shared with English
- Includes all words that look similar to English words that share a similar meaning

Example 1: {A1 example}

---------------------------------------------------
Prompt:


Write a short story (3-5 paragraphs) with the following plot. Output the story only and no other text!

Plot: {story plot}    
\end{lstlisting}

\subsection{Supervised finetuning and PPO}

For supervised finetuning, we used the HuggingFace library \cite{hf-wolf-etal-2020-transformers} to train with the causal language modelling objective. We used the Adam optimizer with beta1=0.9, beta2=0.999. We restricted the maximum sequence length for training to be 4096 tokens. We trained with a weight decay of 1e-2 and a learning rate of 1e-4. For memory efficiency, we used Parameter Efficient Finetuning (PEFT) via QLORA \cite{qlora2023,hf-peft} with $8$-bit quantization and a batch size of $2$. The model was trained on four A6000 GPUs. The LORA parameters were r=16, lora\_alpha=32, and lora\_dropout=0.1.

For the Proximal Policy Optimization, we used the negative of the $\CS{}$ between the generated text and the target level as a reward for the algorithm. We trained using the TRL library \cite{vonwerra2022trl} with adaptive KL penalty, with a KL coefficient of 0.2. We clipped rewards to a clip range of 0.2 and used reward scaling as well as reward normalization. We trained with a learning rate of 1e-5 and also used the same QLORA configuration as the finetuning models for efficiency. 

For generation, we use a standard probabilistic sampling approach with nucleus sampling and top\_k. The parameters for these were as follows: top\_k = 50, top\_p = 0.95, and temperature = 0.7. We limited generation to a maximum length of 2048 tokens. 

Full training details and scripts are included with our code release.

\section{Choosing an Automatic CEFR Scorer}\label{appx:scorer_disc}

Automatically assessing the proficiency level of text is a natural task, but comes with several challenges. A key difficulty with CEFR scoring is the inherent ambiguity in the levels. As can be seen by the official descriptions in \Cref{table:cefr_levels}, each level is coarsely defined with lots of room for interpretation. This makes having a single, correct measure of proficiency difficult.

To understand this ambiguity, we look at three different datasets of CEFR levelled text: the CambridgeExams (CE) dataset of \citet{xia-etal-2016-text}, the EDIA data from the European Language Grid \cite{Breuker2023}, and a dataset compiled on Kaggle for different texts \cite{montgomerieCEFRdata}. The first two of these are gold-standard, in the sense that they are labelled by human experts.

\begin{figure}[h]
    \centering
    \includegraphics[width=0.5\linewidth]{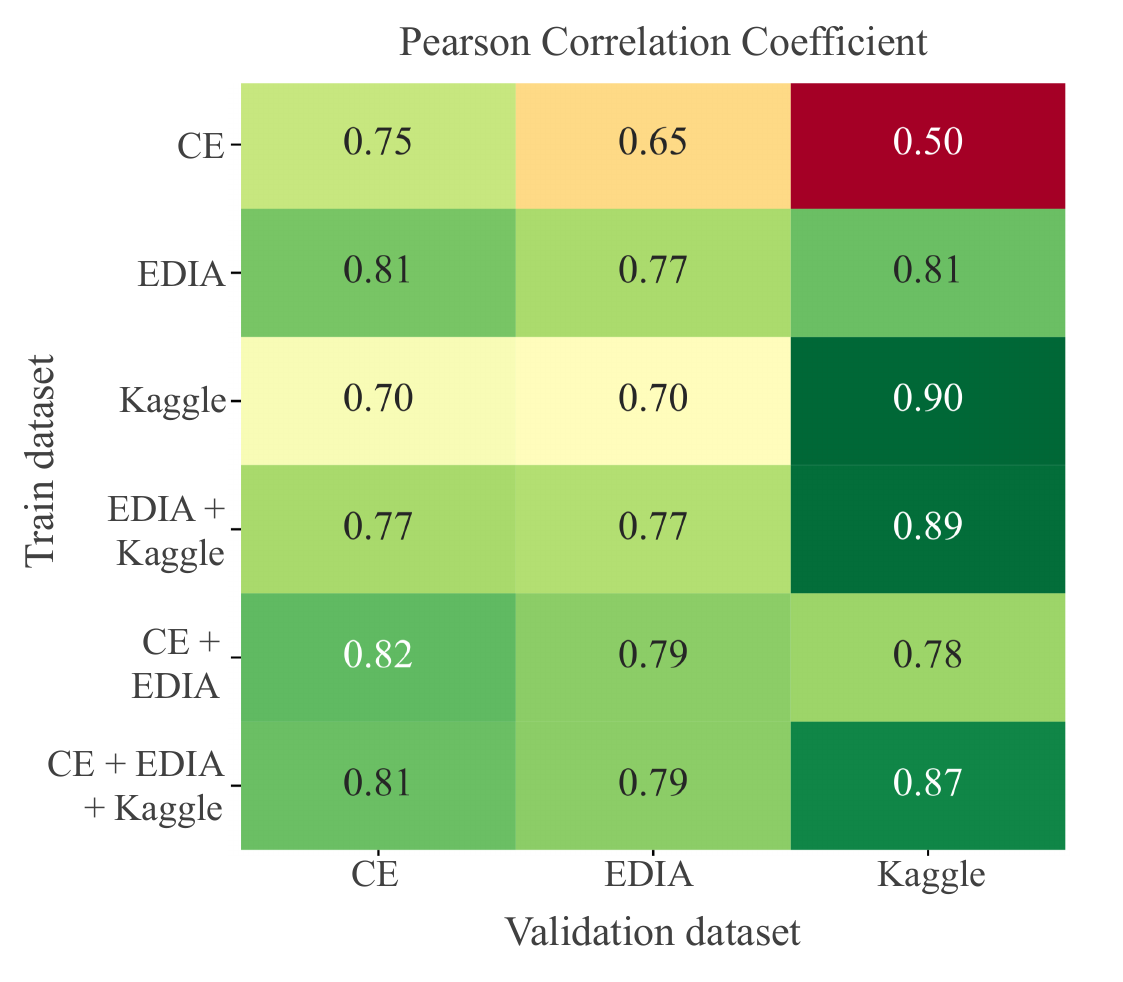}
    \caption{\textbf{Different CEFR datasets introduce distribution shift. } Pearson correlation coefficient of predictions made by a CEFR scorer trained on a particular dataset and evaluated on another. Performance drops off the diagonal due to distribution shift.}
    \label{fig:cefr_datasets}
\end{figure}

We can measure the generalisation capability of scoring functions trained on one of these datasets and evaluated on the others. For example, we train a scorer on CE, and evaluate the Pearson Correlation Coefficient (PCC) of its predictions on CE, EDIA, and Kaggle. We use PCC instead of something more direct because the ability to compare texts in an ordinal sense has shown to be a better measure of generalisability in CEFR scoring \cite{xia-etal-2016-text}. We also look at training on mixtures of datasets.  \Cref{fig:cefr_datasets} shows the results for each different training dataset evaluated on all the others. We see a clear differentiation within each dataset, with no single one performing well on the other two. This is likely due to the inherent differences in interpretation of CEFR levels in the labelling process. We do unsuprisingly find that a mixture of datasets generalises best. 

\begin{figure}[h]
    \centering
    \includegraphics[width=0.5\linewidth]{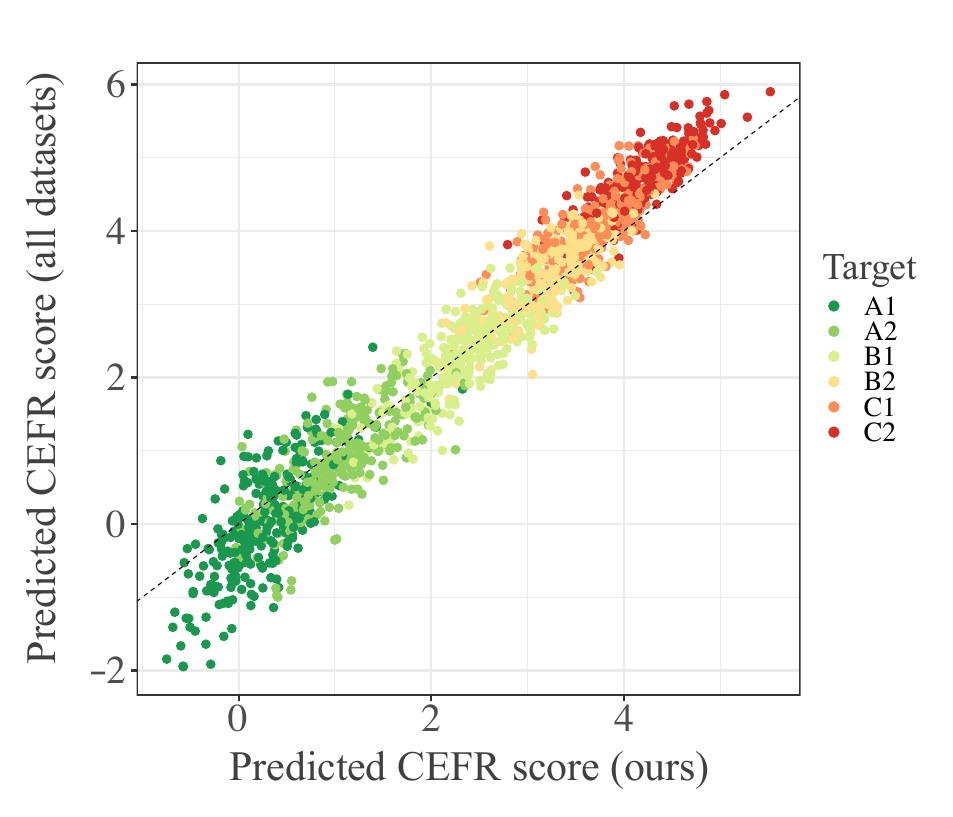}
    \caption{\textbf{CEFR scoring functions with different dataset choices are largely comparable.} Comparison between our scoring function and a different scorer trained on a mixture of all datasets (CambridgeExams + EDIA + Kaggle). Scores evaluated on all data generated from \Cref{tab:prompt_results}, Column 1.}
    \label{fig:cefr_scorers}
\end{figure}

At the time of running our experiments, we only had access to the EDIA and Kaggle data. Thus the scoring function in the experiments is trained on a mixture of the two. Nevertheless, we believe the results in this paper, such as the fact that GPT4 outperforms open source models and that open source models can match GPT4's performance with finetuning and PPO, hold for any reasonable scoring function we could have used.

Some concrete evidence of the robustness of our results to different scoring functions can be seen in \Cref{fig:cefr_scorers}, which shows the relationship between scores predicted by our scoring function and the arguably ``stronger'' one trained on a mixture of all three datasets. This functions are evaluated on the text generation by GPT4 prompting strategies in \Cref{tab:prompt_results}, Column 1. We see that the scores are highly correlated, with a Pearson correlation coefficient of 0.977 ($p=0$).

\section{Human Evaluation Instructions}
\label{sec:eval-instructions}
The following are the instructions we gave for the human evaluations.

``Each row consists of a story plot prompt and an AI generated story. The generated story should follow the plot of the prompt and be written in correct English.
Your goal is to evaluate the generated story on two criteria:

\textbf{Consistency} (scale of 1 to 5): This measures how consistent the generated story is with the plot summary in the prompt. In other words, does the summary accurately characterise the story?
\begin{itemize}
    \item 5: Means the story perfectly follows the plot
    \item 4: The story mostly follows the plot, with a few minor detail differences such as character names or objects
    \item 3: The story roughly follows the plot but there are notable inconcistencies with the plot
    \item 2: The story hardly follows the plot, mostly ignoring it or going off into a different direction.
    \item 1: The story has nothing to do with the plot.
\end{itemize}

\textbf{Language Score} (scale of 1 to 5): This measures how fluent the story is in terms of correct use of English grammar and sentences. It does NOT measure how complex or proficient the writing is.
\begin{itemize}
    \item 5: Perfect use of English. The writing is natural and has no mistakes
    \item 4: The text is perfectly written but might have some slightly awkward phrases.
    \item 3: The text is pretty good but has a few minor grammar mistakes.
    \item 2: The text has a lot of mistakes.
    \item 1: The text is hardly in English
\end{itemize}

Notes:
\begin{itemize}
    \item The stories will vary in writing level from simple, beginner English to advanced writing.
    \item Some stories might not be completed. Just assume the text would continue and make your assessment on the text you can see
\end{itemize}

\section{TinyTolkien Examples}\label{appx:tinytolkien}



\begin{table*}[t]
\centering
\small
    \begin{tabular}{c | p{0.85\linewidth}}
    \toprule
    \textit{Prompt} & \textit{Pandy and his friends play a game called "match" in the forest. A small mouse named Tiny joins in and surprises them by finding a match, teaching them not to judge someone by their size.} \\
        \\
    A1 & Pandy and his friends like to play. They play a game. It is called "match". They play in the forest. They look for things. They find two things that are the same. That is a match. 

    \quad One day, a small mouse comes. His name is Tiny. Tiny wants to play. Pandy and his friends laugh. They say, "You are small. You can't play." But Tiny says he can. He wants to try.
    
    \quad Tiny starts to look. He looks here. He looks there. Then he sees something. He finds a match! Pandy and his friends are surprised. They see Tiny can play. They learn a lesson. They learn not to judge someone by their size.
    \\
    \\
    A2 & 
    Pandy and his friends were big, strong bears. They loved to play a game called "Match" in the forest. In this game, they had to find a pair of things that look the same. It could be leaves, flowers, or stones. One day, while they were playing, a small mouse named Tiny came to them. He wanted to play "Match" too. Pandy and his friends laughed a little. They thought Tiny was too small to play their game. But Tiny was not scared. He wanted to play.

    \quad So, Tiny started to look for a match. He looked under the leaves. He looked in the grass. He even climbed up a tree to look. After a while, Tiny found something. He found two acorns that looked the same. He brought them to Pandy and his friends. They were surprised. They did not think Tiny could find a match.
    
    \quad Pandy and his friends learned a lesson that day. They learned that someone's size does not matter. What matters is their heart and their mind. Tiny was small, but he was smart and brave. He taught them not to judge someone by their size. From then on, Pandy and his friends let Tiny play "Match" with them. And sometimes, Tiny even won the game.
\\
    \midrule
    \textit{Prompt} & \textit{
   Tom takes Sue's shiny staff without permission but later feels guilty and returns it to her, learning a lesson about not taking things that do not belong to him.
    } \\ \\
    A1  & 
    Tom and Sue are good friends. Sue has a shiny staff. It is very nice. Tom loves the staff. One day, Tom takes the staff. But he does not ask Sue. 
    
    \quad Tom is happy with the staff. But then, he feels bad. He thinks, "This is not my staff. I did not ask Sue." He feels guilty. He knows it is not good to take things. 
    
    \quad So, Tom gives the staff back to Sue. He says, "I am sorry, Sue. I took your staff. I did not ask." Sue is happy to have her staff again. Tom learns a lesson. He learns not to take things that are not his. He says, "I will not do it again."
    \\
    \\
     C2  & 
    In the dusky twilight of a warm summer evening, Tom's eyes were captivated by the radiant allure of Sue's enchanting staff, an object of mystic charm and radiant brilliance. The lustrous depth of its gemstones, the intricate carvings in its polished mahogany, and the mesmerising aura of power it emanated were irresistible to him. Despite the insidious whisperings of his conscience, a compelling desire overpowered his restraint, and he surreptitiously appropriated the staff, ensnaring himself in a predicament of guilt and dishonesty.
    
    \quad Tormented by guilt and the irrepressible weight of his clandestine possession, Tom found himself unable to savour the forbidden beauty of the stolen artefact. He was instead consumed by an uneasiness that pervaded his every waking moment, a relentless reminder of his moral failings. His nights were fraught with tumultuous dreams of retribution, his days a haze of remorse. The once coveted staff seemed to metamorphose into a relentless accuser, its glowing aura becoming a flickering emblem of his wrongdoings, a constant reminder of the virtue he had forsaken.
    
    \quad  As the days wore on, the gnawing guilt ate away at Tom, prompting him to contemplate the gravity of his actions. He recognised that he had not only violated Sue's trust but had also compromised his integrity for a transient thrill. The haunting realisation of his moral indiscretion was a turning point, prompting him to rectify his mistake.
    
    \quad Harnessing the courage to face his guilt, Tom returned the staff to Sue, his heart heavy with regret but also relieved at the prospect of redemption. His confession exposed the raw vulnerability he felt, his voice laden with remorse. As Sue accepted the staff, her gaze held a silent reproach but also, surprisingly, a hint of forgiveness that warmed Tom's guilt-ridden soul. This encounter was not just about returning a staff; it was a lesson steeped in values of honesty, integrity, and respect for the belongings of others.
    
    \quad From then on, Tom carried with him the lesson of his transgression, a guiding beacon in his life. He learned that the allure of material possessions could never outweigh the cost of compromised integrity. The staff, once a symbol of undeserved possession, became a timeless reminder of the valuable lesson he learned: to never take what does not rightfully belong to him.
  \\
    \bottomrule
    \end{tabular}
\end{table*}

\begin{table*}
\centering
\small
    \begin{tabular}{c | p{0.85\linewidth}}
    \toprule
    \textit{Prompt} & \textit{
    Tim helps a lost duck find its family in the park and learns the importance of helping others.  
    }
    \\
    \\
    B2 & 
One sunny afternoon, Tim was strolling through the local park when he spotted a tiny, fluffy duckling, seemingly lost and quacking frantically. It was a pitiful sight, and Tim couldn't ignore the little creature's distress. He observed the vicinity, hoping to find the duckling's family, but they were nowhere in sight. His heart went out to the small duckling, now shivering and looking even more frightened.

\quad Tim knew he had to help. Being an animal lover, he couldn't bear the thought of leaving the duckling alone in the park. He decided to pick it up gently and search for the rest of its family. He walked around, holding the duckling close to him, scanning the park's ponds and paths. Finally, after what felt like hours, he spotted a family of ducks near a pond on the other side of the park. Relief washed over him as he approached the family. As he neared, the mother duck squawked loudly, and the duckling in his arms quacked in response.

\quad With a gentle release, Tim set the little duckling down and watched as it waddled quickly towards its family. The mother duck and the duckling nuzzled each other, and Tim could spot the joy and relief in their eyes. The whole experience left Tim with a warm feeling inside. It wasn't just about finding a lost duckling; it was about the sense of fulfillment he experienced through the simple act of helping. From then on, Tim became more conscious about helping others, realizing that even small acts of kindness could bring about significant changes.
    \\ \\ 
    C1 & 
Tim was a creature of habit, his daily routine punctuated by a quiet solace he found in walking through the local park. One sunny afternoon, his solitude was broken by an unusual sight - a small, distressed duckling, seemingly lost and quaking nervously. The sight of the vulnerable creature aroused an instinctive feeling of compassion within Tim. He knelt down, gently cradling the frightened bird in his hands, his eyes scanning the vicinity for any sign of a duck family.

\quad Traversing the park, Tim carried the little creature, weaving through the familiar paths, his keen eyes searching for a gaggle of ducks. He could feel the duckling's heart beating against his palm, it's quacks growing quieter, as if it was reassured by Tim's comforting presence. He found himself talking gently to the duckling, a sense of care manifesting in his actions that was usually reserved for humans. Yet, he found the experience strangely comforting, the simplicity of his mission reminding him of childhood stories where good always triumphed.

\quad Finally, by the tranquil lake nestled in the heart of the park, he spotted them - a fleet of ducks, their feathers reflecting the crystalline blue of the water. He carefully laid the duckling near the water's edge, watching as it tentatively waddled forward and joined the group. The other ducks accepted the little one immediately, and it was a heartwarming scene as the duckling reunited with its family.

\quad As Tim resumed his quiet walk, he felt a newfound sense of warmth and satisfaction. He had assisted a creature in need, and the sight of the reunited duck family left an indelible mark on his heart. It was a simple act, but one that had taught him the immeasurable value of lending a helping hand to those in need. Through the distressed quacks of a tiny duckling, he had learnt an invaluable lesson - the importance of empathy, compassion, and the unassuming power of doing good. He walked on, a smile playing on his lips, his heart light with the joy of having made a difference, however small, in someone's life.  \\ 
\midrule
\textit{Prompt} & \textit{
A noisy train has a dream about a wreck and decides to slow down and be more careful. It meets another noisy train and convinces it to do the same, and they become good friends.
}
\\ \\
A1 & 
There is a train. It is a noisy train. It is fast, too fast. One night, the train has a dream. It dreams about a big wreck. The dream is scary. The train wakes up. It is not happy about the dream. The train thinks, "I should slow down. I should be more careful."

\quad So, the train slows down. It is not fast anymore. It is careful. The train is happier now. It is not scared anymore. The train likes being slow and careful. 

\quad One day, the train meets another train. This train is also very noisy. This train is also very fast. The first train says, "You should slow down. You should be more careful. I had a dream about a big wreck. It was very scary." The second train listens. It thinks about the words. It decides to slow down and be more careful too.

\quad The two trains become friends. They are not noisy anymore. They are not fast anymore. They are slow and careful. They are happy. They like being friends. The end.
\\ \\ 
C2 & 
Under the sable expanse of the night sky, a thunderous steam engine named Prometheus churned relentlessly along its iron tracks, its rhythmic clattering echoing through the neighboring hills. With each lurch forward, Prometheus embodied the audacious spirit of human invention, careening through the dark with the audacity of a fiery comet. However, one fateful night, it had a somnolent vision - a horrifying dream of a catastrophic collision, the twisted wreckage and the muffled silence of an aftermath that gnawed at its mechanical heart. This terrifying foreshadowing ignited a spark of caution in Prometheus, compelling it to restrain its habitual gusto and adopt a more measured pace, mindful of the potential for disaster.

\quad Days spiralled into weeks, and Prometheus' change in demeanour was palpable. Its once tempestuous clatter had mellowed into a congenial hum, its pace reduced to a steady, cautious trot. This transformation did not go unnoticed, and during one of its cautious traverses, it chanced upon Hephaestus, another robust locomotive renowned for its brash velocities. Hephaestus, intrigued by the change in Prometheus, asked for the reason behind this newfound restraint. 

\quad Prometheus, with a sense of earnest urgency, recounted its dire dream, painting an explicit image of the potential horror of an unbridled pursuit of speed. The graphic vision was enough to trigger a reflection within Hephaestus, who, after a moment of profound silence, agreed to follow in Prometheus' tracks. From then on, the two mighty engines adopted a more careful, mindful approach, their thunderous roars now a harmonious symphony echoing through the valleys.

\quad The transformation turned the once competitive nature of their relationship into camaraderie. They continued their traverses across the continent, their journeys punctuated by shared anecdotes and laughter that filled the air. The two locomotives, once symbols of reckless speed, had now become icons of careful diligence, their friendship serving as a beacon of unity amidst the clatter of industry... \\ 
    \bottomrule
    \end{tabular}
\end{table*}

\begin{table*}
\centering
\small
    \begin{tabular}{c | p{0.85\linewidth}}
    \toprule
    \textit{Prompt} & \textit{
    Lucy turns a difficult knot in her shoelace into a pretty bow using her creativity, and impresses her mom.
    }
    \\
    \\
    A1 & 
Lucy is a little girl. She has a problem. Her shoe has a hard knot. She can't open it. She is sad. She wants to wear her shoes.

\quad Lucy thinks. She has an idea. She is good at art. She likes to draw and make things. She uses her creativity. She makes the hard knot into a bow. Now, her shoe has a pretty bow. She is happy.

\quad Lucy's mom comes. She sees the pretty bow. She is surprised. She asks, "Lucy, how did you do it?" Lucy smiles. She says, "I used my creativity!" Mom is impressed. She is proud of Lucy. She says, "Good job, Lucy!"
\\ \\ 
B1 & 
Lucy, a cheerful 10-year-old girl, had a problem. She was trying to tie her shoelaces, but there was a difficult knot that she could not undo. She pulled and tugged, but the knot only seemed to get tighter. Lucy felt frustrated, she wanted to go out and play, but this shoelace was stopping her.

\quad Suddenly, Lucy had a spark of creativity. Instead of trying to undo the knot, why not use it? She thought about it for a moment and then started to work. She carefully began to shape the knot into a loop. With some pushing and pulling, she managed to make a second loop and twisted them together. After a few minutes, she had turned the troublesome knot into a pretty bow. She felt a sense of achievement.

\quad Her mom walked into the room and noticed Lucy's shoes. "Wow, Lucy!" she exclaimed. "That's a beautiful bow! How did you do it?" Lucy smiled and explained her creative solution. Her mom was very impressed and gave her a big hug. "Well done, Lucy," she said. "You turned a difficult situation into something beautiful. I'm so proud of you!" Lucy beamed with joy, proud of her accomplishment. Now, she was ready to go out and play. \\
\midrule
\textit{Prompt} & \textit{
An ambulance with a loud siren helps a sick man who fell down by taking him to the doctor.
} \\ \\
A1 &   
There was a man. His name was Tom. One day, Tom was walking. He was feeling very sick. Then, he fell down. 

\quad People saw Tom fall. They were scared. They did not know what to do. A girl named Lisa saw this. She was smart. She knew what to do. Lisa called for an ambulance.

\quad The ambulance came very fast. It had a loud siren. The siren was saying, "Move! Let us help!" Tom was put into the ambulance. The ambulance people were nice. They wanted to help Tom.

\quad The ambulance took Tom to the hospital. The doctor was there. He checked Tom. He gave him medicine. Tom felt better.

\quad Everyone was happy. Tom was not sick anymore. The ambulance and the doctor helped him. They were heroes. 
\\ \\
C2 & 
In the labyrinth of a bustling metropolis, the piercing wail of an ambulance siren cleaved through the cacophony of urban noise. The ambulance, a beacon of resolute hope amidst the pandemonium, manoeuvred deftly through the congested arteries of the city, its oscillating lights casting urgent, ephemeral shadows on the rain-slicked asphalt. Inside, a dedicated team of paramedics was focused on the challenge at hand: stabilising a man of advanced age who had precipitously collapsed in his home.

\quad The man, a septuagenarian named Albert, had been beset by an unanticipated malaise, his vitality draining away like sand through the fingers of time. Once robust and hale, he now found himself a prisoner within his own frail body, convulsing on the cold floor of his antiquated residence. His neighbour, a vigilant woman known for her alert ears and keen sense of community duty, had been the one to raise the alarm, her frantic 911 call acting as the catalyst for the ongoing medical operation.

\quad In the ambulance, the paramedics worked with precise choreography, their movements dictated by years of training and experience. They defibrillated the life back into Albert's faltering heart, administered oxygen to coax his gasping lungs into normal rhythm, and set an intravenous line to replenish his depleting fluids. Their actions were a harmonious ballet of medical expertise, executed with the singular goal of preserving life.

\quad Upon arrival at the hospital, a team of health professionals, armed with an arsenal of advanced medical technology, stood prepared for his arrival. The doctor, a stoic figure with a countenance as steady as his hands, awaited Albert, ready to wage war against the silent enemy that threatened to claim his life. As the ambulance doors swung open and Albert was wheeled into the stark, sterile environment of the emergency room, it was evident that while the first battle had been won, the war was only just beginning.

\quad In the face of the relentless adversary that is human mortality, the ambulance served as an intrepid vessel of salvation, ferrying the beleaguered Albert from the precipice of his downfall to the promise of medical intervention. An ardent symphony of sirens, it served as the clarion call that rallied the forces of life against the specter of death. Through the endeavours of unsung heroes and the relentless pursuit of medical science, Albert was granted a fighting chance, his story a testament to the indomitable. \\
\bottomrule
\end{tabular}
\end{table*}

\end{document}